\documentclass{article}

\usepackage[preprint]{corl_2022} %

\usepackage[numbers]{natbib}
\usepackage{multicol}
\usepackage{microtype}
\usepackage{xspace}
\usepackage{graphicx}
\usepackage{booktabs,colortbl,tabularx}
\usepackage{graphicx}
\usepackage{bbm}
\usepackage{amsmath}
\usepackage{url}
\usepackage{balance}
\usepackage[subtle]{savetrees}
\usepackage{amssymb}
\usepackage{mathtools}
\usepackage{color}
\usepackage{tabularx}
\usepackage{algorithm}
\usepackage[noend]{algorithmic}
\usepackage{subfigure}
\usepackage{enumitem}
\usepackage[export]{adjustbox}

\usepackage{titlesec}
\titlespacing\section{0pt}{3pt plus 4pt minus 2pt}{0pt plus 2pt minus 2pt}
\titlespacing\subsection{0pt}{3pt plus 4pt minus 2pt}{0pt plus 2pt minus 2pt}
\titlespacing\subsubsection{0pt}{3pt plus 4pt minus 2pt}{0pt plus 2pt minus 2pt}

\newcommand{\algname}{ROSIE\xspace} %
\newcommand{\fullalgname}{\textbf{Ro}bot Learning with \textbf{S}emantically \textbf{I}magened \textbf{E}xperience \textbf{(ROSIE)}}

\newcommand{\fullalgnamenocap}{Robot Learning with Semantically Imagened Experience (ROSIE)}

\newcommand{\im}{Imagen\xspace}
\newcommand{\ime}{Imagen Editor\xspace}
\usepackage{xcolor}

\definecolor{llm}{HTML}{5A8FEE}
\definecolor{grounded}{HTML}{E84A3D}
\definecolor{combined}{HTML}{A354F0}	
\usepackage{color, colortbl}
\definecolor{lightblue}{HTML}{e6f3ff}
\definecolor{lightorange}{HTML}{fff2e6}

\definecolor{light-gray}{rgb}{0.8, 0.8, 0.8}
\definecolor{comment-green}{rgb}{0.435, 0.576, 0.106}
\definecolor{prompt-blue}{HTML}{2596be}
\definecolor{code-function}{HTML}{379fbe}
\definecolor{code-function}{HTML}{693da8}  %
\definecolor{code-syntax}{HTML}{0060b1}
\definecolor{code-constant}{HTML}{d86001}
\definecolor{prompt-gray}{HTML}{757575}
\definecolor{highlight}{HTML}{f8f9cb}
\definecolor{highlight}{HTML}{e3eeff}  %
\definecolor{code-perception}{HTML}{2ecc71}
\definecolor{code-control}{HTML}{ff9900}
\definecolor{code-undefined}{HTML}{ff0000}
\renewcommand\fbox{\fcolorbox{light-gray}{white}}
 
\newcommand{\hlcode}[1]{\colorbox{highlight}{\makebox[0.99\linewidth][l]{#1}}}

\newcommand{\query}[1]{\textcolor{comment-green}{#1}}

\newcommand\blfootnote[1]{%
  \begingroup
  \renewcommand\thefootnote{}\footnote{#1}%
  \addtocounter{footnote}{-1}%
  \endgroup
}

\begin{document}

\title{Scaling Robot Learning with \\Semantically Imagined Experience}

\author{
  \textbf{Tianhe Yu$^1$, Ted Xiao$^1$, Austin Stone$^1$, Jonathan Tompson$^1$, } \\
  \textbf{Anthony Brohan$^1$, Su Wang$^2$, Jaspiar Singh$^1$, Clayton Tan$^1$, Dee M$^1$, } \\
  \textbf{Jodilyn Peralta$^1$, Brian Ichter$^1$, Karol Hausman$^1$, Fei Xia$^1$} \\
  
  $^1$Robotics at Google, $^2$Google Research \\
  Project website: \href{https://diffusion-rosie.github.io}{https://diffusion-rosie.github.io}
 
}

\blfootnote{Correspond to \url{{tianheyu, xiafei}@google.com}.}

\maketitle

\begin{abstract}
Recent advances in robot learning have shown promise in enabling robots to perform a variety of manipulation tasks and generalize to novel scenarios. 
One of the key contributing factors to this progress is the scale of robot data used to train the models.
To obtain large-scale datasets, prior approaches have relied on either demonstrations requiring high human involvement or engineering-heavy autonomous data collection schemes, both of which are challenging to scale. 
To mitigate this issue, we propose an alternative route and leverage text-to-image foundation models widely used in computer vision and natural language processing to obtain meaningful data for robot learning without requiring additional robot data. We term our method \fullalgname.
Specifically, we make use of the state of the art text-to-image diffusion models and perform aggressive data augmentation on top of our existing robotic manipulation datasets via inpainting various unseen objects for manipulation, backgrounds, and distractors with text guidance. 
Through extensive real-world experiments, we show that manipulation policies trained on data augmented this way are able to solve completely unseen tasks with new objects and can behave more robustly w.r.t. novel distractors. In addition, we find that we can improve the robustness and generalization of high-level robot learning tasks such as success detection through training with the diffusion-based data augmentation.
\end{abstract}

\section{Introduction}
\label{sec:intro}

Though recent progress in robotic learning has shown the ability to learn a number of language-conditioned tasks~\cite{jiang2022vima, brohan2022rt, shridhar2022cliport, shridhar2022perceiver}, the generalization properties of such policies is still far less than that of recent large-scale vision-language models~\cite{ramesh-dalle2,saharia2022photorealistic,chen2022pali}.
One of the fundamental reasons for these limitations is the lack of diverse data that covers not only a large variety of motor skills, but also a variety of objects and visual domains. 
This becomes apparent by observing more recent trends in robot learning research -- when scaled to larger, more diverse datasets, current robotic learning algorithms have demonstrated promising signs towards more robust and performant robotic systems~\cite{jiang2022vima,brohan2022rt}. 
However, this promise comes with an arduous challenge: it is difficult to significantly scale up diverse, real-world data collected by robots as it requires either engineering-heavy autonomous schemes such as scripted policies~\citep{kalashnikov2021mt,lee2021beyond} or laborious human teleoperations~\citep{jang2022bc,brohan2022rt}. To put it into perspective, it took 17 months and 13 robots to collect 130k demonstrations in~\cite{brohan2022rt}.  In~\cite{kalashnikov2021mt}, the authors used 7 robots and 16 months to collect 800k autonomous episodes.
While some works~\cite{xia2018gibson, kolve2017ai2, savva2019habitat} have proposed potential solutions to this conundrum by generating simulated data to satisfy these robot data needs, they come with their own set of challenges such as generating diverse and accurate enough simulations~\cite{jiang2022vima} or solving sim-to-real transfer~\cite{mehta2020active,sadeghi2017sim2real}. Can we find other ways to synthetically generate realistic diverse data without requiring realistic simulations or data collection on real robots?

To investigate this question we look to the field of computer vision. 
\begin{figure}[t]
     \centering
     \includegraphics[trim={0 0 0 0},clip,width=0.7\linewidth]{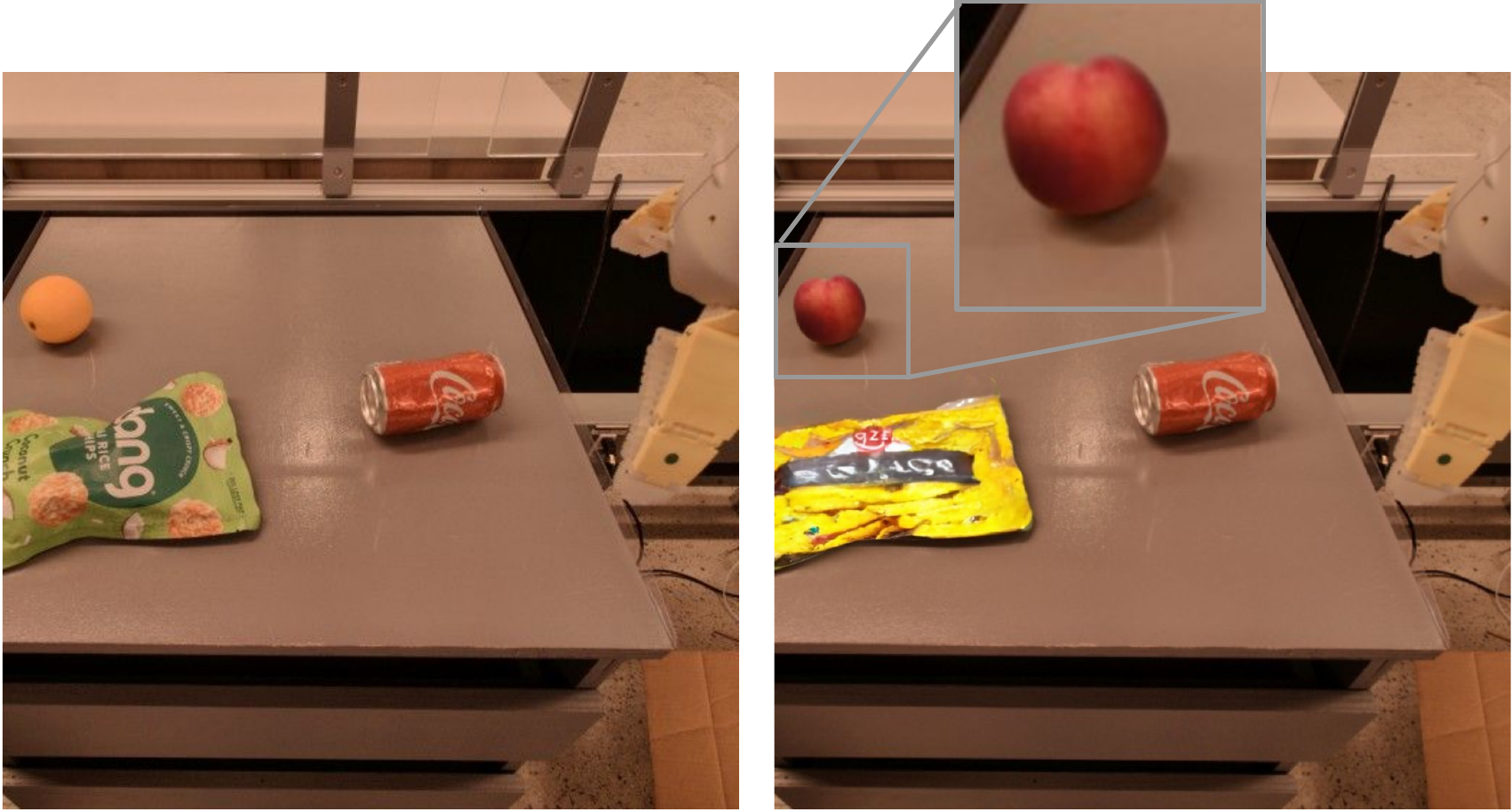}
     \caption{We propose using text-guided diffusion models for data augmentation within the sphere of robot learning. These augmentations can produce highly convincing images suitable for learning downstream tasks. As demonstrated in the figure, some of the objects were produced using our system, and it is difficult to identify which are real and which are generated due to the photorealism of our system.
     }\label{fig:teaser}
     \vspace{-0.5cm}
\end{figure}
Traditionally, synthetic generation of additional data, whether to improve the accuracy or robustify a machine learning model, has been addressed through data augmentation techniques. 
These commonly include randomly perturbing the images including cropping, flipping, adding noise, augmenting colors or changing brightness. 
While effective in some computer vision applications, these data augmentation strategies do not suffice to provide novel robotic experiences that can result in a robot mastering a new skill or generalizing to semantically new environments~\cite{sadeghi2017sim2real,akkaya2019solving,laskin2020curl}.
However, recent progress in high-quality text-to-image diffusion models such as DALL-E 2~\cite{ramesh-dalle2}, Imagen~\cite{saharia2022photorealistic} or StableDiffusion~\cite{rombach-cvpr-2022} provides a new level of data augmentation capability.
Such diffusion-based image-generation methods allow us to move beyond traditional data augmentation techniques, for three reasons.
First, they can meaningfully augment the semantic aspects of the robotic task through a natural language interface.
Second, these methods are built on internet-scale data and thus can be used zero-shot to generate photorealistic images of many objects and backgrounds.
Third, they have the capability to meaningfully change only part of the image using methods such as inpainting~\cite{yu2018generative}.
These capabilities allow us to generate realistic scenes by incorporating novel distractors, backgrounds, and environments while reflecting the semantics of the new task or scene -- essentially distilling the vast knowledge of large generative vision models into robot experience.

As an example, given data for a task such as ``move the green chip bag near the orange'', we may want to teach the robot to move the chip bag of any colors near many new objects that it has not interacted with, such as ``move the yellow chip bag near the peach'' (Fig ~\ref{fig:teaser}). 
These techniques allow us to exchange the objects from real data for arbitrary relevant objects.
Furthermore, they can leave the semantically relevant part of the scene untouched, e.g. the grasp of the chip bag remains, while the orange becomes a peach.
This results in a novel, semantically-labelled data point to teach the model a new task.
Such a technique can reasonably generate many more examples such as ``move the apple near the orange on a wooden desk'', ``move the plum near the orange'', or even ``place the coke can in the sink''.

In this paper, we investigate how off-the-shelf image-generation methods can vastly expand robot capabilities, enabling new tasks and robust performance.
We propose \fullalgname, a general and semantically-aware data augmentation strategy.
\algname works by first parsing human provided novel instructions and identifying areas of the scene to alter.
It then leverages inpainting to make the necessary alterations, while leaving the rest of the image untouched.
This amounts to a \textit{free lunch} of novel tasks, distractors, semantically meaningful backgrounds, and more, as generated by internet-scale-trained generative models.
We demonstrate this approach on a large dataset of robotic data and show how a subsequently trained policy is able to perform novel, unseen tasks, and becomes more robust to distractors and backgrounds. Moreover, we show that \algname\ can also improve the robustness of success detection in robotic learning especially in out-of-distribution (OOD) scenarios.

\section{Related Work}
\noindent\textbf{Scaling robot learning.}
Given the recent results on scaling data and models in other fields of AI such as language~\cite{brown2020language,devlin2018bert,chowdhery2022palm} and vision~\cite{dosovitskiy2020image,alayrac2022flamingo,chen2022pali}, there are multiple approaches trying to do the same in the field of robot learning.
One group of methods focuses on scaling up robotic data via simulation~\cite{jiang2022vima,shridhar2022perceiver,shridhar2020alfred,savva2019habitat,james2020rlbench,yu2020meta,mittal2023orbit,xiang2020sapien} with the hopes that the resulting policies and methods will transfer to the real world. The other direction focuses on collecting large diverse datasets in the real world by either teleoperating robots~\cite{mandlekar2019scaling,jang2022bc,brohan2022rt,ebert2021bridge} or autonomously collecting data via reinforcement learning~\cite{kalashnikov2021mt,kalashnikov2018qtopt,lee2021beyond} or scripting behaviors~\cite{dasari2019robonet}. 
In this work, we present a complementary view on scaling the robot data by making use of state-of-the-art text-conditioned image generation models to enable new robot capabilities, tasks and more robust performance.

\noindent\textbf{Data augmentation and domain randomization.} Domain randomization~\cite{mehta2020active, tobin2017domain, tremblay2018training} is a common technique for training machine learning models on synthetically generated data. The advantage of domain randomization is that it makes it possible to train models on a wide variety of data to improve generalization. 
Domain randomization usually involves changing the physical parameters or rendering parameters (lighting, texture, backgrounds) in simulation models~\cite{laskin2020reinforcement,kostrikov2020image,hansen2020self,li2021domain}. Others use data augmentation to transformer simulated data to be more realistic~\cite{sadeghi2017sim2real,akkaya2019solving,rao2020rl,ho2021retinagan} or vice-versa~\cite{james2019sim}.
Contrary to these methods, we propose to directly augment data collected in the real world. We operate directly on the real-world data and leverage diffusion models to perform photorealistic image manipulation on this data.

\noindent\textbf{Diffusion models for robot control.}
Though diffusion models~\citep{sohl2015deep,ho2020denoising,song2020score,song2020improved,song2020denoising,nichol2021improved,dhariwal2021diffusion,ho2022classifier,saharia2022photorealistic,ramesh-dalle2} have become common-place in computer vision, their application to robotic domains is relatively nascent. \citet{janner2022planning} uses diffusion models to generate motion plans in robot behavior synthesis.
Some works have used the ability of image diffusion models to generate images and perform common sense geometric reasoning to propose goal images fed to object-conditioned policies~ \cite{liu2022structdiffusion,kapelyukh2022dall}.
The recent concurrent works CACTI~\cite{mandi2022cacti} and GenAug~\citep{chen2023genaug} are most similar to ours. CACTI proposes to use diffusion model for augmenting data collected from the real world via adding new distractors and requires manually provided masks and semantic labels. GenAug explores the usage of depth-guided diffusion models for augmenting new tasks and objects in real-world robotic data with human-specified masks and object meshes.
In contrast, our work generates both novel distractors and new tasks without requiring depth. In addition, it \emph{automatically} selects regions for inpainting with text guidance and leverages text-guided diffusion models to generate novel, realistic augmentations.

\section{Preliminaries}
\label{sec:prelim}

\noindent\textbf{Diffusion models and inpainting.}
Diffusion models are a class of generative models that have shown remarkable success in modeling complex distributions~\cite{sohl2015deep}.
Diffusion models work through an iterative denoising process, transforming Gaussian noise into samples of the distribution guided by a mean squared error loss.
Many such models also have the capability for high-quality \textit{inpainting}, essentially filling in masked areas of an image~\cite{efros2001image,pathak2016context,iizuka2017globally,yu2018generative}.
In addition, such approaches can be guided by language, thus generating areas consistent with both a language prompt and the image as a whole~\cite{wang2022imagen}.

\noindent\textbf{Multi-task language-conditioned robot learning.} 
Herein we learn vision and language-conditioned robot policies via imitation learning.
We denote a dataset $\mathcal{D} \coloneqq \{\mathbf{e}_j\}_{j=1}^N$ of $N$ episodes $\mathbf{e} = \{(\mathbf{o}_i, \mathbf{a}_i, \mathbf{o}_{i+1}, \ell)\}_{i=1}^T$ where $\mathbf{o}$ denotes the observation, which correspond to the image in our setting, $\mathbf{a}$ denotes the action, and $\ell$ denotes the language instruction of the episode, identifying the target task. 
We then learn a policy $\pi(\cdot | \mathbf{o}_i, \ell)$ to generate an action distribution by minimizing the negative-log liklihood of actions, i.e. \textit{behavioral cloning}~\cite{pomerleau1988alvinn}.
To perform large-scale vision-language robot learning, we train the RT-1 architecture~\cite{brohan2022rt}, which utilizes FiLM-conditioned EfficientNet~\citep{pmlr-v97-tan19a}, a TokenLearner~\citep{ryoo2021tokenlearner}, and a Transformer~\citep{vaswani2017attention} to output actions.
\section{\fullalgnamenocap}

\begin{figure*}[t]
     \centering
     \includegraphics[trim={0 0 0 0},clip,width=1.0\textwidth]{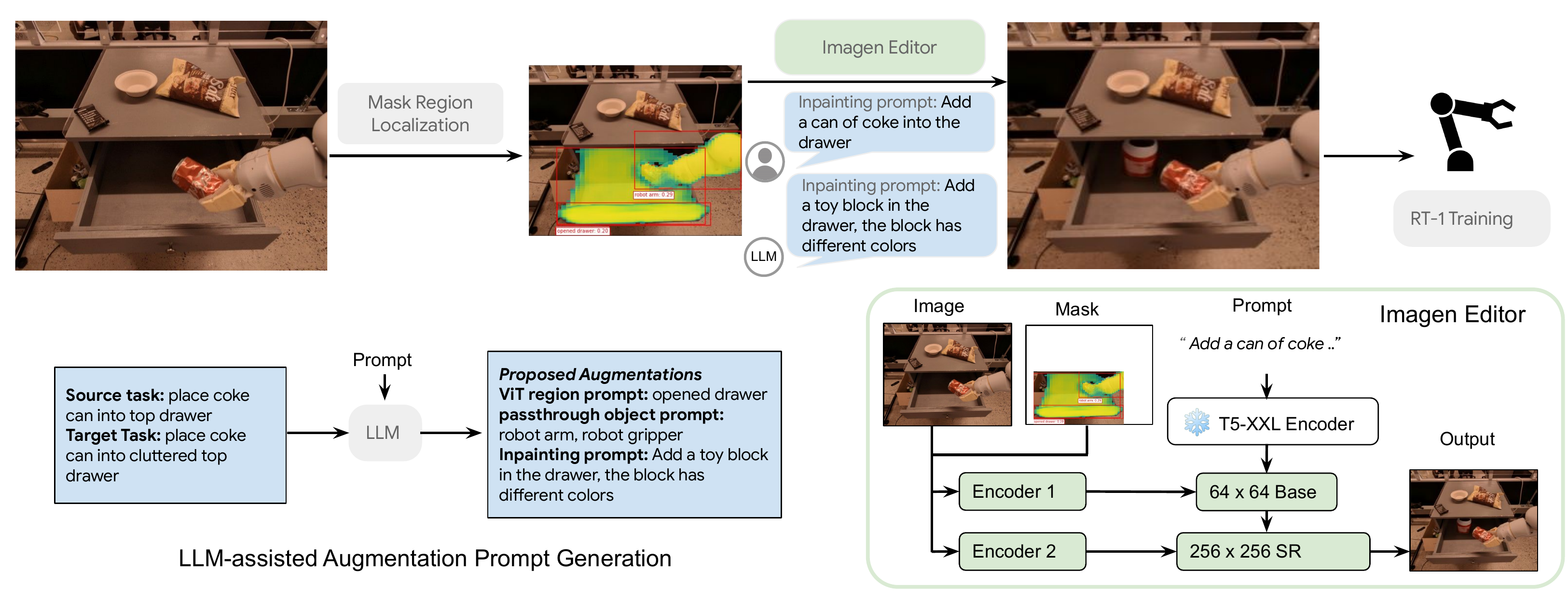}
     \caption{The proposed architecture of \algname. First, we localize the augmentation region with open vocabulary segmentation model. Second, we run Imagen Editor to perform text-guided image editing. Finally, we use the augmented data to train an RT-1 manipulation policy~\cite{brohan2022rt}. Concretely, we explain \algname\ using the example shown in the figure as follows. We take the original episode with the instruction ``place coke can into top drawer'' and the goal is to add distractors in the opened drawer to improve the robustness of the policy. For \textit{each image} in the episode, we detect the masks of the open drawer, the robot arm, and the coke can using our first step. We obtain the mask of the target region to add the distractor via subtracting the masks of the robot arm and the coke can that is picked up from the mask of the open drawer. Then, we generate our augmentation proposal leveraging LLMs as described in Section~\ref{sec:aug_text}. We run \ime{} with the augmentation text and the selected mask to generate a coke can in the drawer discussed in Section~\ref{sec:diffusion_inpaint}. We combine both the original episodes and the augmented episodes and perform policy training using multi-task imitation learning.
     }\label{fig:system_architecture}
\end{figure*}

In this section, we introduce our approach, \algname, an automated pipeline for scaling up robot data generation via semantic image augmentation.
We assume that we have access to episodes of state and action pairs demonstrating a robot executing a task that is labelled with a natural language instruction.
As the first step of the pipeline we augment the natural language instruction with a semantically different circumstance.
For example, given a demonstration of placing an object in an empty drawer, we add ``there is a coke can in the opened drawer''.
With this natural language prompt, \algname generates the mask of the region of interest that is relevant to the language query.
Next, given the augmentation text, \algname performs inpainting on the selected mask with Imagen Editor~\cite{wang2022imagen} to insert semantically accurate objects that follow the augmented text instruction. 
Importantly, the entire process is applied throughout the robot trajectory, which is now consistently augmented across all the time steps. We present the overview of this pipeline in Fig.~\ref{fig:system_architecture}.
We describe the details of each component of \algname in the following sections. 
In Section~\ref{sec:segmentation}, we show how we obtain the mask of the target region using open vocabulary segmentation. In Section~\ref{sec:aug_text}, we discuss two main approaches to proposing prompts used for Imagen Editor, which can be either specified manually or generated automatically with a large language model. In Section~\ref{sec:diffusion_inpaint}, we discuss how we perform inpainting with Imagen Editor based on the augmentation prompt. Finally, we show how we use the generated data in downstream tasks such as policy learning and learning high-level tasks such as success detection in Section~\ref{sec:policy_learning}.

\begin{figure}[h]
     \centering
     \includegraphics[trim={0 0 0 0},clip,width=0.85\linewidth]{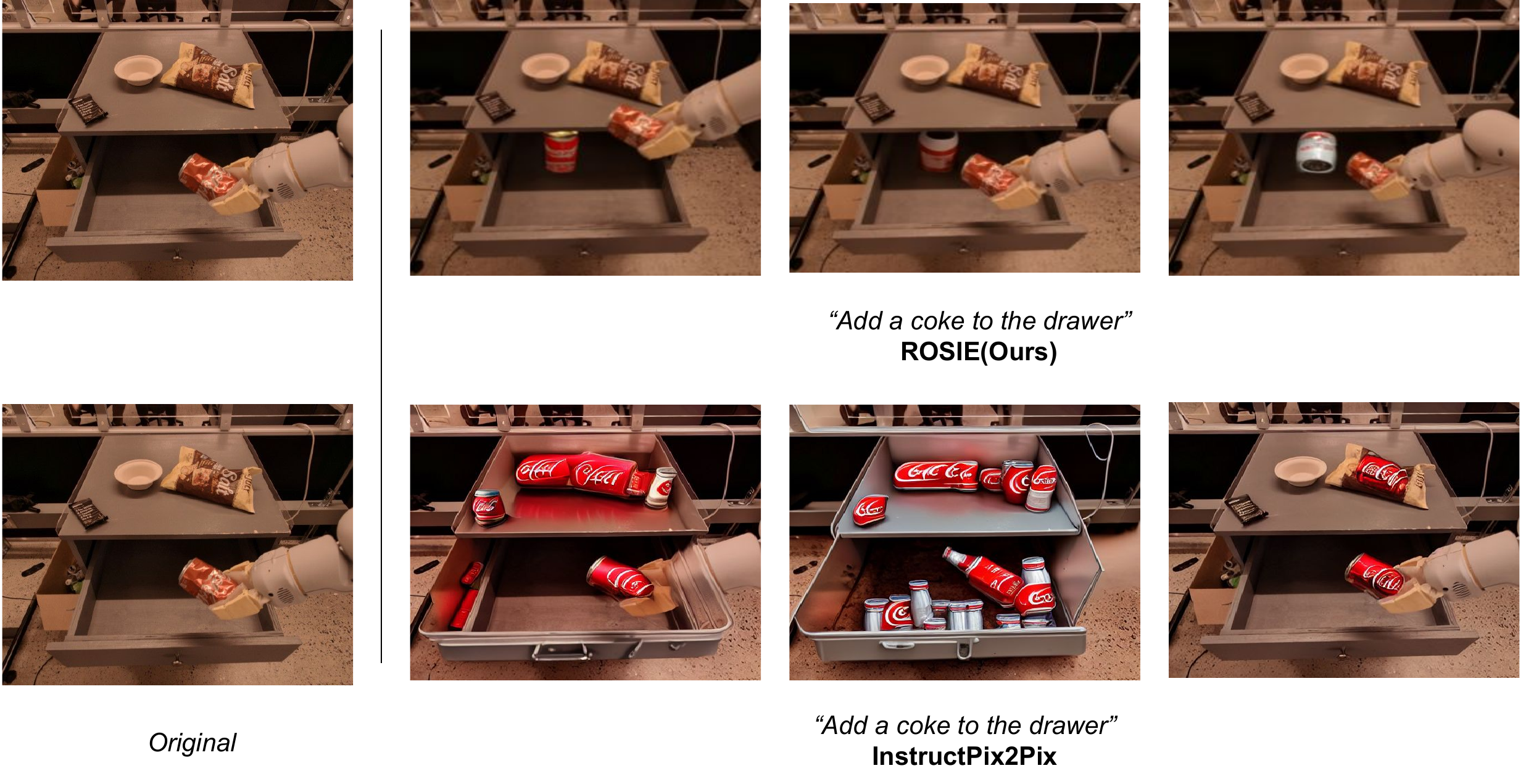}
     \vspace{-0.3cm}
     \caption{Our augmentation scheme generates more targeted and physically realistic augmentations that are useful for learning downstream tasks, while other text-to-image generation methods such as InstructPix2Pix~\cite{brooks2022instructpix2pix} often makes global changes rendering the image unusable for training.
     }\label{fig:qualitative}
     \vspace{-0.5cm}
\end{figure}

\begin{figure}[h]
     \centering
     \includegraphics[trim={0 0 0 0},clip,width=0.8\linewidth]{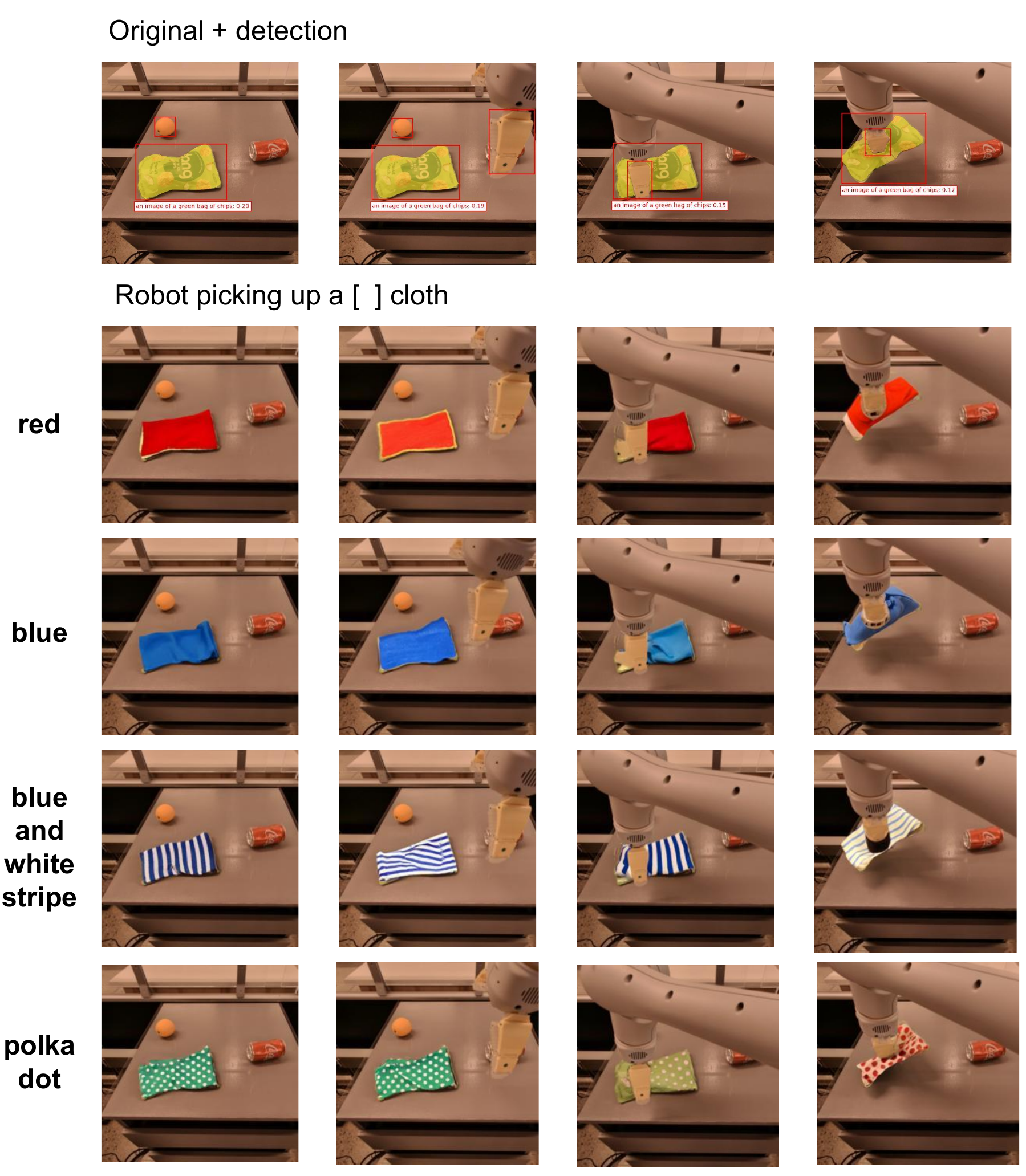}
     \caption{Augmentations of in-hand objects during manipulation. We show examples where \algname\ effectively inpaint novel objects into the original in-hand objects during manipulation. On the top row, we show the original episode with detected masks where the robot picks up the green chip bag. On the following row, we show that \algname\ can inpaint various microfiber cloth with different colors and styles into the original green chip bag. For example, we can simply pass the original episode with the masks and the prompt \texttt{Robot picking up a polka dot cloth} to get an episode the robot picking such cloth in a photorealistic manner.
     }\label{fig:qualitative2}
     \vspace{-0.7cm}
\end{figure}

\subsection{Augmentation Region Localization using Open Vocabulary Segmentation}
\label{sec:segmentation}

In order to generate semantically meaningful augmentations on top of existing robotic datasets, we first need to detect the region of the image where such augmentation should be performed. To this end, we perform open-vocabulary instance segmentation leveraging the OWL-ViT open-vocabulary detector~\cite{minderer2022simple} with an additional instance segmentation head. 
This additional head predicts fixed resolution instance masks for each bounding box detected by OWL-ViT (similar in style to Mask-RCNN \cite{maskrcnn}). 
In particular, we freeze the main OWL-ViT model and fine-tune a mask head on Open-Images-V5 instance segmentations~\cite{OpenImages,OpenImagesSegmentation}.

\begin{figure}[t]
     \centering
     \includegraphics[trim={0 0 0 0},clip,width=0.65\linewidth]{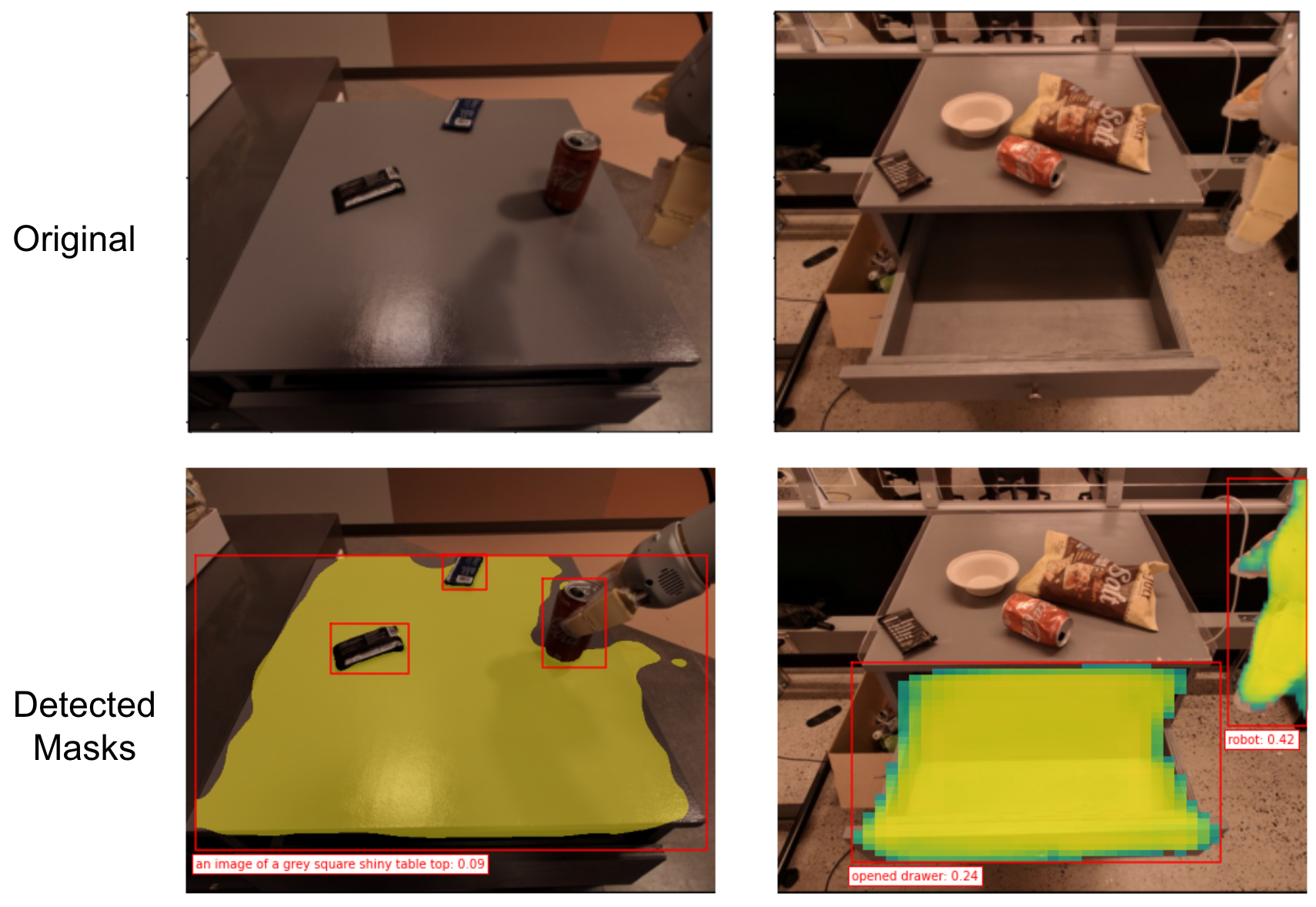}
     \caption{We show the original images from RT-1 datasets on the top row and the images with detected masks and mask labels on the bottom row.
     }\label{fig:detected_masks}
     \vspace{-0.3cm}
\end{figure}

The instance segmentation model provided by OWL-ViT requires a language query that specifies which part of the image should be detected. 
We can generate masks for objects that the robot arm interacts with. Given each episode $\mathbf{e}$ in our robotic dataset, we first identify the target objects specified in the language instruction $\ell$. For example, if $\ell$ is ``pick coke can'', the target object of the task is a coke can.
We pass the target object as a prompt to the OWL-ViT model to perform segmentation and obtain the resulting mask. 
We can also generate masks in regions where distractors can be inpainted to improve the robustness of policy. In this setting, we use the OWL-ViT to detect both the table (shown in Figure~\ref{fig:system_architecture}) and all the objects on the table. 
This allows us to sample a mask on the table in a way that it does not overlap with existing objects (which we call passthrough objects). We provide more examples of masks detected by OWL-ViT from our robotic dataset in Figure~\ref{fig:detected_masks}.

\subsection{Augmentation Text Proposal}
\label{sec:aug_text}

Next, we discuss two main approaches to attain the augmentation prompt for the text-to-image diffusion model: hand-engineered prompt and LLM-proposed prompt.

\textbf{Hand-engineered prompt.} The first method involves manually specifying the object to augment. For generating new tasks, we choose objects that lie outside of our training data to ensure that the augmentations are able to expand the data support. For improving robustness of the learned policy and success detection, we randomly pick objects that are semantically meaningful and add them in the prompt to generate meaningful distractors in the scene. For example, in Figure~\ref{fig:qualitative2} where we aim to generate novel in-hand objects by replacing the original object (green chip bag) with various microfiber cloth, we use the following prompt \texttt{Robot picking up a blue and white stripe cloth} to effectively perform inpainting.

\begin{figure*}[t]
     \centering
     \subfigure[]{
     \includegraphics[trim={0 0 0 0},clip,width=\linewidth]{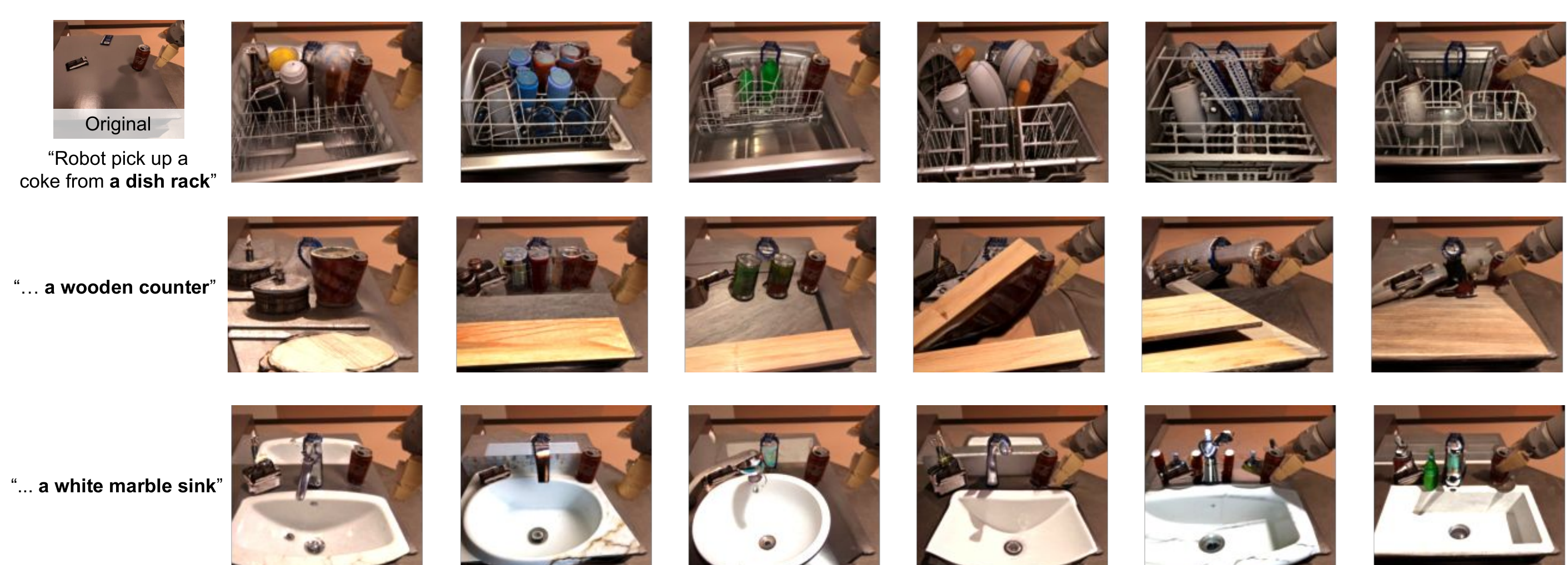}
     }
     \subfigure[]{
     \includegraphics[trim={0 0 0 0},clip,width=\linewidth]{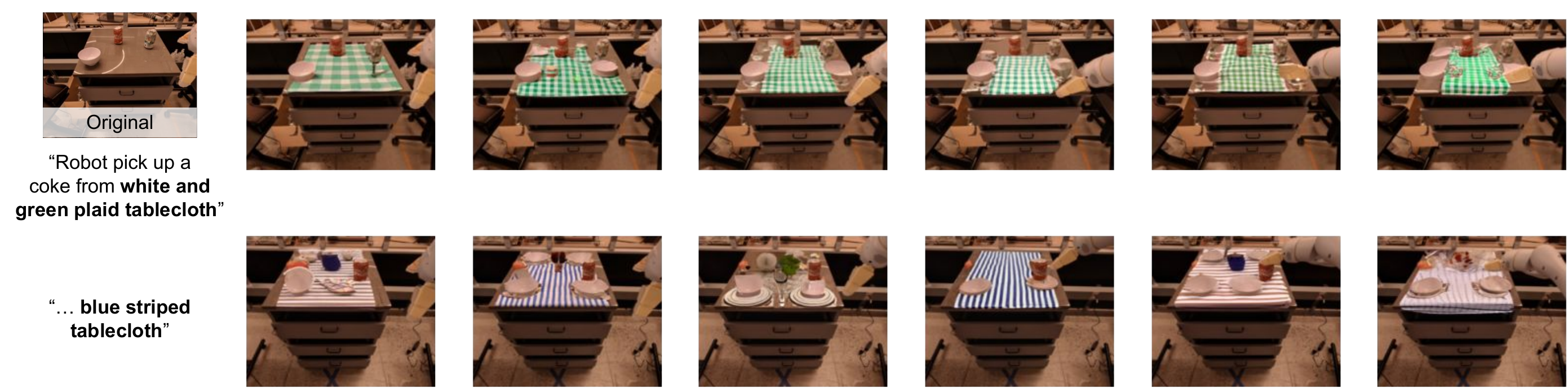}
     }
     \subfigure[]{
     \includegraphics[trim={0 0 0 0},clip,width=\linewidth]{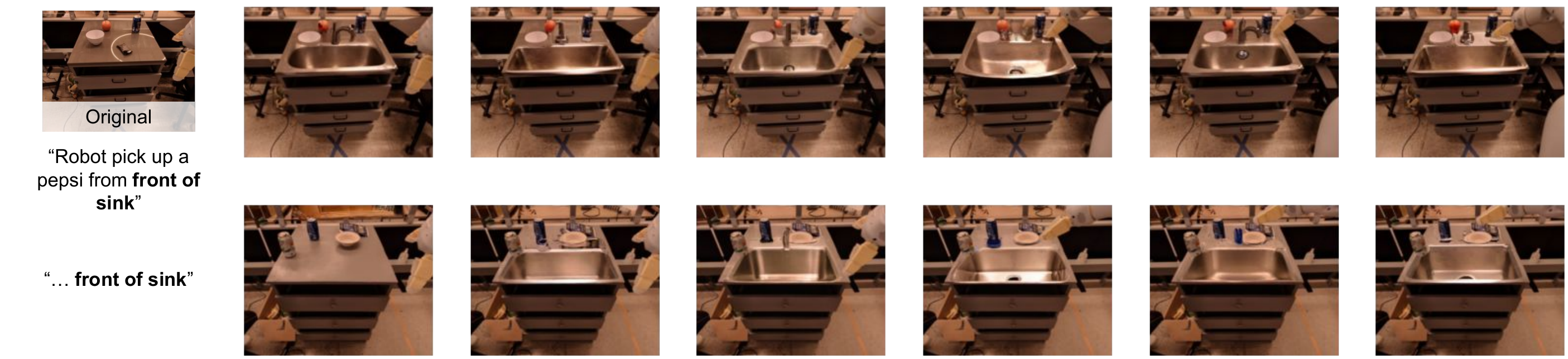}
     }
     \vspace{-0.2cm}
     \caption{We show visualizations of the episodes generated by \algname where we replace the regular tabletop in front of the robot with a dish rack, a marble sink and a wooden counter, which never appears in the training dataset. Our results in Section~\ref{subsec:new_skills} and Figure~\ref{fig:sink_eval} show that the policy trained on such augmentations enables the robot to place objects into a real metal sink.
     }\label{fig:aug_example4}
     \vspace{-0.6cm}
\end{figure*}

\textbf{LLM-proposed prompt.} While hand-engineered prompt may guarantee the generated data to be out-of-distribution, it makes the data generation process less scalable. Therefore, we propose to leverage the power of large language models in proposing objects to augment. We leverage the rich semantics learned in LLMs to propose a vast list of objects with detailed descriptions of visual features for augmentation. 
We employ GPT-3~\citep{brown2020language} as our choice of LLM to propose the augmentation text. 
In particular,  we specify the original task of the episode and the target task after augmentation in the LLM prompt, and ask the LLM to propose the OWL-ViT prompt for detecting masks of both the target region and the passthrough objects. 
We present an example of LLM-assisted augmentation prompt proposal in Figure~\ref{fig:system_architecture}, where LLM-generated augmentation text is highly informative, which in turn benefits the text-guided image editing. Therefore, we use LLM-proposed prompts in our experiments. Despite that there is some noise in the LLM-proposed prompts (see Appendix~\ref{app:failure_cases}), it generally does not hurt robotic control performance in practice.

\subsection{Diffusion Model for Text-Guided Inpainting}
\label{sec:diffusion_inpaint}

Given the segmentation mask and the augmentation prompt, we perform text-guided image editing via a text-to-image diffusion model.
Herein, we use Imagen Editor~\citep{wang2022imagen}, the latest state-of-the-art text-guided image inpainting model fine-tuned on pre-trained text-to-image generator \im~\citep{saharia2022photorealistic}, though we note that our approach, \algname, is agnostic to the choice of inpainting models. 
\ime~\citep{wang2022imagen} is a cascaded diffusion architecture. All of the diffusion models, i.e., the base model and super-resolution (SR) models (i.e., conditioned on high-resolution 1024$\times$1024 image and mask inputs) are trained with new convolutional image encoders shown in the bottom right corner of Figure~\ref{fig:system_architecture}.
\ime is capable of generating high-resolution photorealistic augmentations, which is crucial for robot learning as it relies on realistic images capturing physical interactions.
Moreover, \ime{} is trained to de-noise object-oriented masks provided by off-the-shelf object detectors~\citep{ssd-mobilenet-v2} along with random box/stroke masks~\citep{lama-suvorov2021resolution}, enabling inpainting with our mask generation procedure.

To summarize more formally, given a robotic episode $\mathbf{e} = \{(\mathbf{o}_i, \mathbf{a}_i, \mathbf{o}_{i+1}, \ell)\}_{i=1}^T$, the mask $\mathbf{m}$ designating the target area(s) to be modified, and our generated augmentation text $\ell_\text{aug}$, we iteratively query \ime{} with input $\mathbf{o}_i$, $\mathbf{m}$ and $\ell_\text{aug}$ over $i=1, \dots, T$. 
As a result, \ime{} generates the masked region according to the input text $\ell_\text{aug}$ (e.g. inserting novel objects or distractors) while ensuring consistency with the unmasked and unedited content of $\mathbf{o}_i$.
This results in generating augmented image $\tilde{\mathbf{o}}_i$.
In scenarios where $\ell_\text{aug}$ creates a new task, we modify the instruction $\ell$ to $\tilde{\ell}$. For example, as shown in Figure~\ref{fig:qualitative2} where we replace the green chip bag with various styles of microfiber cloth, we modify the original instruction $\ell=$ ``pick green rice chip bag'' to $\tilde{\ell}=$ ``pick blue microfiber cloth'', pick ``polka dot microfiber cloth'' and etc.
The actions $\{\mathbf{a}_i\}_{i=1}^T$ remain unchanged, as \ime{} alters novel objects consistently with the semantics of overall image.
In summary, \algname\ eventually yields the augmented episode $\tilde{\mathbf{e}} = \{(\tilde{\mathbf{o}_i}, \mathbf{a}_i, \tilde{\mathbf{o}}_{i+1}, \tilde{\ell})\}_{i=1}^T$. Powered by the expressiveness of diffusion models and priors learned from internet-scale data, \algname\ is able to provide physically realistic augmentations (e.g. Figure~\ref{fig:qualitative}) that are valuable in making robot learning more generalizable and robust, which we will show in Section~\ref{sec:exps}.

\subsection{Manipulation Model Training}
\label{sec:policy_learning}

The goal of the augmentation is to improve learning of downstream tasks, e.g. robot manipulation. We train a manipulation policy based on Robotics Transformer (RT-1) architecture~\cite{brohan2022rt} discussed in Section~\ref{sec:prelim}.
Given the \algname augmented dataset $\mathcal{\tilde{D}} \coloneqq \{\tilde{\mathbf{e}}_{j}\}_{j=1}^{\tilde{N}}$, where $\tilde{N}$ is the number of augmented episodes, we train a policy on top of a pre-trained RT-1 model~\citep{brohan2022rt}.
The finetuning uses a 1:1 mixing ratio of $\mathcal{D}$ and $\mathcal{\tilde{D}}$. 
We follow the same training procedure described in~\citep{brohan2022rt} except that we use a smaller learning rate $1\times10^{-6}$ to ensure the stability of fine-tuning.

\section{Experiments}
\label{sec:exps}

In our experimental evaluation, we focus on robot manipulation and embodied reasoning (e.g. detecting if a manipulation task is performed successfully). We design experiments to answer the following research questions:

\begin{figure*}[t]
     \centering
     \includegraphics[trim={0 0 0 0},clip,width=\linewidth]{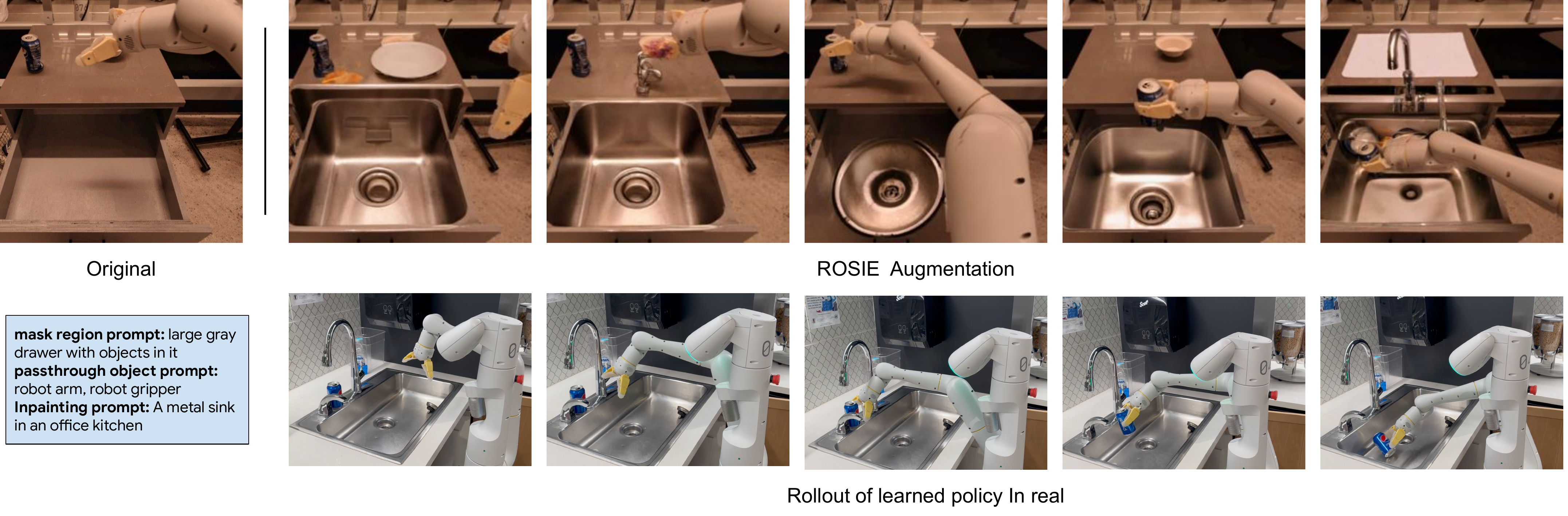}
     \vspace{-0.3cm}
     \caption{We show an episode augmented by \algname\ (top row) where \algname\ inpaints the metal sink onto the top drawer of the counter and a rollout of policy trained with both the original episodes and the augmented episodes in a real kitchen with a metal sink. The policy successfully performs the task ``place pepsi can into sink'' even if it is not trained on real data with sink before, suggesting that leveraging the prior of the diffusion models trained with internet-scale data is able to improve generalization of robotic learning in the real world.
     }\label{fig:sink_eval}
     \vspace{-0.5cm}
\end{figure*}

\begin{enumerate}
    \item \textbf{RQ1}: Can we leverage semantic-aware augmentation to learn completely new skills only seen through diffusion models?
    \item \textbf{RQ2}: Can we leverage semantic-aware augmentation to make our policy more robust to visual distractors?
    \item \textbf{RQ3}: Can we leverage semantic-aware augmentation to bootstrap high-level embodied reasoning such as success detection?
\end{enumerate}
To answer these questions, we perform empirical evaluations of \algname\ using the multi-task robotic dataset collected in \citep{brohan2022rt}, which consists of $\sim$130k robot demonstrations with 744 language instructions collected in laboratory offices and kitchens. These tasks include skills such as picking, placing, opening and closing drawers, moving objects near target containers, manipulating objects into or out of the drawers, and rearranging objects. For more details regarding the tasks and the data used we refer to~\citet{brohan2022rt}.

In our experiments, we aim to understand the effects of both the augmented text and the augmented images on policy learning.
We thus perform two comparisons, ablating these changes:
\begin{enumerate}
    \item \textbf{Pre-trained RT-1 (NoAug)}: we take the RT-1 policy trained on the 744 tasks in \citep{brohan2022rt}. While pre-trained RT-1 is not trained on tasks with the augmentation text and generated objects, it has been shown to enjoy promising pre-training capability and demonstrate excellent zero-shot generalization to unseen scenarios \citep{brohan2022rt} and therefore, should have the ability to tackle the novel tasks to some extent.
    \item \textbf{Fine-tuned RT-1 with Instruction Augmentation (InstructionAug)}: Similar to ~\citet{xiao2022robotic}, we relabel the original episodes in RT-1 dataset to new instructions generated via our augmentation text proposal~\ref{sec:aug_text} while keeping the images unchanged. We expect this method to bring the text instructions in-distribution but fail to recognize the visuals of the augmented objects.
\end{enumerate}

For implementation details and hyperparameters, please see Appendix~\ref{app:details}.

\subsection{RQ1: Learning new skills}
\label{subsec:new_skills}

To answer RQ1, we augment the RT-1 dataset via generating new objects that the robot needs to manipulate. We evaluate our method and the baselines in the following four categories with increasing level of difficulty.

\paragraph{\textbf{Learning to move objects near generated novel containers}} First, we test the tasks of moving training objects near unseen containers. We visualize such unseen containers in Figure~\ref{fig:aug_example3} in Appendix~\ref{app:aug_examples}. We select the tasks ``move \{some object\} near white bowl'' and ``move \{some object\} near paper bowl'' within the RT-1 dataset, which yields 254 episodes in total. We use the augmentation text proposals to replace the white bowl and the paper bowl with the following list of objects \{lunch box, woven basket, ceramic pot, glass mason jar, orange paper plate\}, which are visualized in Figure~\ref{fig:aug_example3}. 
For each augmentation, we augment the same number of episodes as the original task. 

As shown in Table~\ref{tab:ravens_full}, our \algname fine-tuned RT-1 policy (trained on both the whole RT-1 training set of 130k episodes and the generated novel tasks) outperforms pre-trained RT-1 policy and fine-tuned RT-1 with instruction augmentations, suggesting that \algname\ is able to generate fully unseen tasks that are beneficial for control and exceeds the inherent transfer ability of RT-1.

\paragraph{\textbf{Learning to place objects into generated unseen containers}} Second, we perform a similar experiment, where we focus on \textit{placing} objects into the novel target containers, rather than just nearby.
Example augmentations are shown in Figure~\ref{fig:aug_example3}.
Table~\ref{tab:ravens_full} again shows \algname\ outperforms both pre-trained RT-1 and RT-1 with instruction augmentation by at least 75\%.

\paragraph{\textbf{Learning to grasp generated unknown deformable objects}} Third, we test the limits of \algname on novel tasks where the object to be manipulated is generated via \algname. We pick the set of tasks ``pick green chip bag'' from the RT-1 dataset consisting of 1309 episodes. To accurately generate the mask of the chip bag throughout the trajectory, we run our open-vocabulary segmentation to detect the chip bag and the robot gripper as the passthrough objects so that we can filter out the robot gripper to obtain the accurate mask of the chip bag when it is grasped. 
We further query \ime{} to substitute the chip bag with a fully unknown microfiber cloth with distinctive colors (black and blue), with augmentations shown in Figure~\ref{fig:qualitative2}. Table~\ref{tab:ravens_full} again demonstrates that \algname outperforms pre-trained RT-1 and RT-1 with instruction augmentation by at least 150\%, proving that \algname is able to expand the manipulation task family via diversifying the manipulation targets and boost the policy performance in the real world.

\paragraph{\textbf{Learning to place objects into an unseen kitchen sink in a new background}} Finally, to further stress-test our diffusion-based augmentation pipeline, we try to learn to place object into a sink. Note that the robot has never collected data for that task in the real world. We generate a challenging scenario where we take the all the RT-1 tasks that perform placing a can into the top drawer of a counter (779 episodes in total) and deploy \algname\ to detect the open drawer and replace the drawer with a metal sink using \ime{} (see the first row of Figure~\ref{fig:sink_eval} for the visualization). Similar to the above two experiments, we dynamically compute the mask of the open drawer at each frame of the episode while removing the robot arm and the can in the robot hand from the mask. Note that the generated sink makes the scene completely out of the training distribution, which poses considerable difficulty to the pre-trained RT-1 policy. The results in the last row in Table~\ref{tab:ravens_full} confirm this. \algname\ achieves 60\% overall success rate in placing the coke can and the pepsi can into the sink whereas the RT-1 policy is not able to locate the can and fails to achieve any success. In Figure~\ref{fig:sink_eval}, we include the visualizations of a trajectory of the original episode with augmentations that replaces the drawer with the sink and a trajectory of the policy rollout performing the task near a real metal sink. Our method effectively learns from the episodes with the sink generated by \algname\ and completes the task that involve the sink in the real kitchen.

Overall, through these experiments, \algname\ is shown to be capable of effectively inpainting both the objects that require rich manipulation and the target object of the manipulation policy, significantly augmenting the number of tasks in robotic manipulation. 
These results indicate a promising path to scaling robot learning without extra effort of real data collection.

\begin{table}[t]
\begin{center}
\setlength\tabcolsep{2.0pt}
\footnotesize
\bgroup
\def\arraystretch{1.2}
\resizebox{0.75\columnwidth}{!}{\begin{tabular}{@{}lcccccc@{}}
    \toprule
      \multicolumn{1}{l}{\textbf{\textcolor{gray}{Task Family}} / Text Instruction} &  \multicolumn{1}{c}{\textbf{NoAug}} & \multicolumn{1}{c}{\textbf{InstructionAug}} &  \multicolumn{1}{c}{\textbf{\algname}} \\
    
    \midrule
    \small{\textbf{\textcolor{gray}{Move object near \textbf{novel} object}}} & 0.86 & 0.78 & \textbf{0.94}  \\
    \rowcolor{lightblue}
    move coke can/orange near lunch box & 0.8 & 0.6 & 0.9\\
    \rowcolor{lightblue}
    move coke can/orange near woven basket & 0.7 & 0.6 & 0.9\\
    \rowcolor{lightblue}
    move coke can/orange near ceramic pot  & 1.0 & 0.9 & 1.0\\
    \rowcolor{lightblue}
    move coke can/orange near glass mason jar & 0.9 & 0.8 & 1.0\\ 
    \rowcolor{lightblue}
    move coke can/orange near orange paper plate & 0.9 & 1.0 & 0.9\\ 
    
    \midrule
    \small{\textbf{\textcolor{gray}{Pick up \textbf{novel} object}}} & 0.25 & 0.3 & \textbf{0.75} \\
    \rowcolor{lightblue}
    pick blue microfiber cloth & 0.1 & 0.4 & 0.8 \\
    \rowcolor{lightblue}
    pick black microfiber cloth & 0.4 & 0.2 & 0.7 \\
   
    \midrule
    \small{\textbf{\textcolor{gray}{Place object into \textbf{novel} container}}} & 0.13 & 0.25 & \textbf{0.44} \\
    \rowcolor{lightblue}
    place coke can into orange plastic plate & 0.0 & 0.19 & 0.5\\
    \rowcolor{lightblue}
    place coke can into blue plastic plate & 0.25 & 0.06 & 0.38\\
    
    \midrule
    \small{\textbf{\textcolor{gray}{Place object into \textbf{sink}}}} & 0.0 & - & \textbf{0.6} \\
    \rowcolor{lightblue}
    place coke can into sink & 0.0 & - & 0.8\\
    \rowcolor{lightblue}
    place pepsi can into sink & 0.0 & - & 0.4\\
    
    \midrule
    \small{\textbf{\textcolor{gray}{Pick up object in new backgrounds}}} & 0.33 & - & \textbf{0.71}  \\
    \rowcolor{lightorange}
    pick coke can on an orange table cloth  & 0.0 & - & 0.4 \\
    \rowcolor{lightorange}
    pick pepsi can on an orange table cloth  & 0.0 & - & 0.7 \\
    \rowcolor{lightorange}
    pick coke can on an blue and white table cloth  & 0.2 & - & 0.7 \\
    \rowcolor{lightorange}
    pick pepsi can on an blue and white table cloth  & 0.8 & - & 0.8 \\
    \rowcolor{lightorange}
    pick coke can near the side of a sink  & 0.4 & - & 0.5 \\
    \rowcolor{lightorange}
    pick pepsi can near the side of a sink  & 0.3 & - & 0.7 \\
    \rowcolor{lightorange}
    pick coke can in front of a sink  & 0.4 & - & 0.9 \\
    \rowcolor{lightorange}
    pick pepsi can in front of a sink  & 0.5 & - & 1.0 \\
    \midrule
    \small{\textbf{\textcolor{gray}{Place object into cluttered drawer}}} & 0.38 & - & \textbf{0.55} \\
    \rowcolor{lightorange}
    place blue chip bag into top drawer & 0.5 & -  & 0.4 \\ 
    \rowcolor{lightorange}
    place green jalapeno chip bag into top drawer & 0.4 & - & 0.5 \\ 
    \rowcolor{lightorange}
    place green rice chip bag into top drawer & 0.4 & - & 0.5 \\ 
    \rowcolor{lightorange}
    place brown chip bag into top drawer & 0.2 & - & 0.8 \\ 
    \midrule
    \small{\textbf{\textcolor{gray}{Pick up object (with OOD distractors)}}} & 0.33 & - & \textbf{0.37}  \\
    \rowcolor{lightorange}
    pick coke can & 0.33 & - & 0.37 \\
    \bottomrule
\end{tabular}}
\egroup
\vspace{0.3cm}
\caption{Full Experimental Results for \algname. The blue shaded results correspond to RQ1 and the orange shaded results correspond to RQ2. For each task family from top to the bottom, we performed evaluations with 50, 20, 16, 10, 80, 40, and 27 episodes respectively (243 episodes in total). \algname\ outperforms \textbf{NoAug} (pre-trained RT-1 policy) and \textbf{InstructionAug} (fine-tuned RT-1 policy with instruction augmentation~\citep{xiao2022robotic}) in both categories, suggesting that \algname\ can significantly improve the generalization to novel tasks and robustness w.r.t. different distractors.}
\label{tab:ravens_full}
\vspace{-0.8cm}
\end{center}
\end{table}

\subsection{RQ2: Robustifying manipulation policies}
\label{sec:robustify}

We investigate RQ2 with two scenarios: policy robustness w.r.t. different backgrounds and new distractors. 

\paragraph{\textbf{Unseen background.}}We employ \algname\ to augment the background in our training data. We perform two types of augmentations: replacing the table top with a colorful table cloth and inserting a sink on the table top. We select two manipulation tasks, ``pick coke can'' and ``pick pepsi can'' from our training set, which consists of 1222 episodes in total. We run open-vocabulary segmentation to detect the table and passthrough objects, which consist of the robot arm and the target can. To generate a diverse set of table cloth during augmentation, we query GPT-3 with the following prompt:
\vspace{0.3em}
\noindent\fbox{\parbox{0.97\linewidth}{\scriptsize{\texttt{{%
{\color{prompt-gray}
inpainting prompt: pick coke can from a red and yellow table cloth}\\
\query{goal: list  30 more table cloth with different vivid colors and styles with visual details}\\
\color{prompt-gray}
inpainting prompt: pick coke can from\\
\hlcode{1. Navy blue and white striped table cloth\\}
\hlcode{2. White and pink polka dot table cloth\\}
\hlcode{3. Mint green and light blue checkered table cloth\\}
\hlcode{4. Cream and gray floral table cloth\\}
\hlcode{5. Hot pink and red floral table cloth\\}
\hlcode{...
}}}}}}\\
We show the some example answers from GPT-3 in blue, which are semantically meaningful. We use \ime{} to replace the table top except the target can with the LLM-proposed table cloth. To inpaint a sink on the table, we follow the same procedure described in the placing objects into unseen sink task in Section~\ref{subsec:new_skills} except that we inpaint the sink on the table top rather than the open drawer. We present visualizations of such augmentations in Figure~\ref{fig:aug_example4}. We fine-tune the pre-trained RT-1 policy on both the original data and the augmented episodes with generated table cloth and metal sink. As shown in Table~\ref{tab:ravens_full}, \algname\ + RT-1 signifcantly outperforms RT-1 \textbf{NoAug} in 7 out of 8 settings while performing similarly to \textbf{NoAug} in the remaining scenario, achieving an overall 115\% improvement. Therefore, \algname\ is highly effectively in robustifying policy performance under varying table textures and background.

\paragraph{\textbf{Novel distractors.}}To test whether \algname\ can improve policy robustness w.r.t. novel distractors and cluttered scenes, we consider the following two tasks. First, we train a policy solely from the task ``pick coke can'' and investigate its ability to perform this task with distractor coke cans, which have not been seen in the 615 training episodes.
To this end, we employ \algname to add an equal number of augmented episodes with additional coke cans on the table (see Figure~\ref{fig:aug_example} in Appendix~\ref{app:aug_examples} for visualizations). 
As shown in Table~\ref{tab:ravens_full}, RT-1 + \algname\ augmentations improves the performance over RT-1 trained with ``pick coke can'' data only in scenarios where there are multiple coke cans on the table.

Second, we evaluate a task that places a chip bag into a drawer and investigate its ability to perform this task with distractor objects already in the drawer, also unseen during training. This scenario is challenging for RT-1, since the distractor object in the drawer will confuse the model and make it more likely to directly output termination action.
We use \algname to add novel objects to the drawer, as shown in Figure~\ref{fig:aug_example2} in Appendix~\ref{app:aug_examples} and follow the same training procedure as in the coke can experiment. 
Table~\ref{tab:ravens_full} shows that RT-1 trained with both the original data and \algname\-generated data outperforms RT-1 with only original data. Our interpretation is that RT-1 trained from the training data never sees this situation before and it incorrectly believes that the task is already solved at the first frame, whereas \algname\ can mitigate this issue via expanding the dataset using generative models.

\subsection{RQ3: A Case Study on Success Detection}
\label{sec:sd}

In this section, we show that \algname\ is also effective in improving high-level robotic embodied reasoning tasks, such as success detection. Success detection (or failure detection) is an important capability for autonomous robots for accomplishing tasks in dynamic situations that may require adaptive feedback from the environment.
Given large diversity of potential situations that a robot might encounter, a general solution to this problem may involve deploying learned failure detection systems~\citep{mtopt2021arxiv} that can improve with more data.
As recent work~\citep{xiao2022robotic} has shown, visual-language models (VLMs) such as CLIP~\citep{radford2021learning} with internet scale pre-training can be fine-tuned on domain specific robotic experience to perform embodied reasoning such as success detection.
However, collecting domain specific fine-tuning data is often expensive, and it is difficult to scale data collection to cover all potential success and failure cases.
This challenge is similar to the one of learning a robust policy that we presented in the previous sections, where the dataset of robot data might include data distribution biases that are difficult to correct with on-robot data collection alone.

As a motivating example, consider the experimental setting from Section~\ref{subsec:new_skills} where a large dataset of teleoperated demonstrations was collected for placing various household objects into empty cabinet drawers. 
A success detector trained on this dataset would require additional priors and/or data to generalize to images of cluttered drawers.

To study this setting, we utilize \algname\ to augment 22764 episodes of placing objects into drawers tasks from the dataset used in \citep{xiao2022robotic} and then fine-tune a CLIP-based success detector following the procedure in \citep{xiao2022robotic}.
Starting from the episodes of robotic placing into empty drawers, we create two augmented datasets with \algname\ to emulate visual clutter: one dataset \textbf{(A)} that includes generated distractor chip bags inside the drawer and one dataset \textbf{(B)} that includes generated soda cans inside the drawer. 
Both datasets have the same number of episodes as the original dataset. We evaluate the fine-tuned CLIP-based success detector with and without \algname-augmented episodes in two datasets: the in-distribution set and the OOD set. Our in-distribution set contains 76 episodes of robot putting green rice chip bag into the drawer and taking it out of the drawer, while the OOD set contains 58 episodes of robot putting (green rice, green japaleno, blue, brown) chip bag into the drawer, but the drawer contains other items, which are not observed in the training set. Note that this OOD set makes success detection particularly challenging as the model can easily be misguided by the cluttered distractors in the drawer and make incorrect predictions even if the robot fails to place the target object into the drawer.

By utilizing increasing amounts of augmentation from \algname, we find that learned success detectors become increasingly robust detecting successes and failures in real-world difficult cluttered OOD drawer scenarios in terms of F1 score, as seen in Table \ref{tab:exp_clip}. Note that our OOD dataset is highly challenging, as discussed above, so that the prior work~\cite{xiao2022robotic} without augmentations struggles a lot in this setting whereas \algname\ obtains a reasonable performance.
Furthermore, we find that the accuracy on the standard, in-distribution tasks remains unchanged.
This indicates that \algname can be used as a general semantically-consistent data augmentation technique across various tasks such as policy learning and embodied reasoning.

\begin{table}[t]
    \centering
    \resizebox{0.6\columnwidth}{!}{\begin{tabular}{lccc}
\toprule
 & No Aug & \algname\ Aug \textbf{(A)} & \algname\ Aug (\textbf{(A)} + \textbf{(B)}) \\
 \midrule
\textbf{Overall} & 0.43 & 0.56 & \textbf{0.62} \\ 
In-Distribution set & 0.66 & \textbf{0.67} & 0.66 \\ 
OOD set & 0.19 & 0.45 & \textbf{0.57} \\ 
 \bottomrule
\end{tabular}}
    \vspace{0.3cm}
    \caption{CLIP success detection Results. \algname\ improves the robustness of the success detection on hard OOD cases as the number of augmentations increases. All numbers are the F1 score and we use 0.5 as the threshold. We augment the data with datasets \textbf{A} and \textbf{B}, which include different distractors as described in text. }
    \label{tab:exp_clip}
    \vspace{-0.8cm}
\end{table}

\section{Societal Impact}
The model used in this work is a text-guided image generation model, which open many new possibilities for content creation and subsequently many risks. 
Our approach attempts to minimize many of these risks through a controlled usage of these technologies, by only modifying local patches of images and using narrowly scoped semantic labels. 
We further follow accepted responsible AI practices,
such as regularly inspecting data before training on it, and in general recommend researchers to establish robust inspection and filtering mechanisms when utilizing text-guided image generation models for data augmentation.

\section{Discussion, Future Work, and Conclusion}

In summary, we presented \algname, a system that uses off-the-shelf text-guided image generation models to vastly expand robotics datasets without any real-world data collection. To accomplish this, we generated new instructions and their corresponding text prompts for alternating the images, enabling robots to achieve tasks that were \emph{only seen through the lens of image generation process}. We were also able to generate semantically meaningful augmentations of the images, enabling various learned models trained on the data to be more robust with respect to OOD scenes. Lastly, we experimentally validated the proposed method on a variety of language-conditioned manipulation tasks.

Though the method is general and flexible, there are a few limitations of this work that we aim to address in the future. First, we only augment the appearance of the objects and scenes, and do not generate new motions. 
To alleviate this limitation of not augmenting physics and motions, we could consider mixing in simulation data as a potential source of diverse motion data.
Another limitation of the proposed method is that it performs image augmentation per frame, which can lead to a loss in temporal consistency. 
However, we find that at least for the architecture that we use (Robotics Transformer~\cite{brohan2022rt}), we do not suffer from a performance drop. State of the art text-to-video diffusion models~\cite{ho2022imagen, singer2022make, villegas2022phenaki, dreamix} can generate temporally consistent videos but might lose photorealism and physics realism. We speculate that this can cause downstream task learning performance to deteriorate. The trade off between photorealism and temporally consistency remains an interesting topic for future studies.
Finally, we use a diffusion model for image augmentation, which is computationally heavy and limits our capability to perform on-the-fly augmentation. As a future direction, we could consider other models such as the mask transformer-based architecture~\cite{chang2023muse}, which is 10x more efficient.

\section*{Acknowledgments}

We would like to acknowledge Sharath Maddineni, Brianna Zitkovich, Vincent Vanhoucke, Kanishka Rao, Quan Vuong, Alex Irpan, Sarah Laszlo, Bob Wei, Sean Kirmani, Pierre Sermanet and the greater
teams at Robotics at Google for their feedback and contributions.

\bibliography{references}

\clearpage
\appendix
\section*{Appendices}
\renewcommand{\thesubsection}{\Alph{subsection}}

\subsection{Experiment Details}
\label{app:details}

\subsubsection{Implementation Details and Hyperparameters}

We take a pre-trained RT-1 policy with 35M parameters and trained for 315k steps at a learning rate of $1\times10^{-4}$ and fine-tune the RT-1 policy with 1:1 mixing ratio of the original 130k episodes of RT-1 data and the \algname-generated episodes with for 85k steps with learning rate $1\times10^{-6}$. We follow all the other policy training hyperparameters used in \cite{brohan2022rt}.

To obtain the accurate segmentation mask of the target region of augmentations, we set a threshold for filtering out predicted masks with low prediction scores of both the region of the interest and passthrough objects given by OWL-ViT. In cases where we have multiple detected masks, we always select the one with highest prediction score. Specifically, for experiments where the robot is required to pick novel objects or place objects into novel containers or move objects near unseen containers (Section~\ref{subsec:new_skills}), we use a threshold of 0.07 to detect the in-hand objects and the containers while using a threshold of 0.05 to detect passthrough objects, which are the robot arm and robot gripper. In experiments where the robot is instructed to place the coke can or the pepsi can into the unknown sink or pick up coke can and the pepsi can with new background , we use a threshold of 0.04 to detect the table with all objects and a threshold of 0.03 to detect the passthrough objects, which are the robot arm, robot gripper and the coke can or the blue can in this case. In experiments discussed in Sections~\ref{sec:robustify} and \ref{sec:sd}, we use the threshold of 0.3 to detect the table or the open drawer where we want to add new distractors.

For generating LLM-assisted prompts, we perform 1-shot prompting to the LLM. For example, in the setting of generating novel distractors in the task where we place objects into the drawer (Section~\ref{sec:robustify}), we use the following prompt to the LLM:

\vspace{0.3em}
\noindent\fbox{\parbox{0.97\linewidth}{\scriptsize{\texttt{{%
{\color{prompt-gray}
Source task: place pepsi can on the counter}\\
{\color{prompt-gray}
Target task: place pepsi can on the clutter counter}\\
\query{ViT region prompt: empty counter}\\
\query{passthrough object prompt: robot arm, robot gripper}\\
\query{inpainting prompt: add a chip bag on the counter}\\
{\color{prompt-gray}
Source task: place coke can into top drawer}\\
{\color{prompt-gray}
Target task: place coke can into cluttered top drawer}\\
}}}}}\\

and LLM generates the following prompt for detecting masks and augmentations (light blue means LLM generated):

\vspace{0.3em}
\noindent\fbox{\parbox{0.97\linewidth}{\scriptsize{\texttt{{%
\hlcode{ViT region prompt: empty drawer\\}
\hlcode{passthrough object prompt: robot arm, robot gripper\\}
\hlcode{inpainting prompt: add a box of crackers in the drawer\\}
}}}}}\\

which is semantically meaningful for performing mask detection and \ime{} augmentation. We follow this recipe of prompting for all of the tasks in our experiments.

During inpainting, we take the checkpoint of \ime{} 64x64 base model and the 256x256 super-resolution model trained in \cite{wang2022imagen} and directly run inference to produce augmentations.

During evaluation, for the tasks that perform moving objects near novel containers and grasping unseen microfiber cloth, we perform 10 policy rollouts per new container/microfiber cloth of each method. For tasks that perform placing objects into novel containers, we perform 8 policy rollouts per new container for each method. For the task where the robot is instructed to place coke can or pepsi can into the unseen kitchen sink, for each method, we perform 5 policy rollouts for coke can and pepsi can respectively. For the task where the robot is instructed to grasp the coke can and the pepsi can in new backgrounds, we evaluate each method with 10 rollouts. For the task where the robot places the object into the cluttered drawer, we perform 10 policy rollouts per object for each method. Finally, for the task that requires the robot to pick up coke can in a scene with multiple coke cans, we perform 27 policy rollouts for each approach.

\subsubsection{Computation Complexity}

We train our policy on 16 TPUs for 1 day. For obtaining segmentation masks, we perform inference of OWL-ViT on 1 TPU for 1 hour to generate 1k episodes. During augmentation, we perform inference of \ime{} using 4 TPUs of the 64 x 64 base model and the 256 x 256 super-resolution model respectively for 2 hours to generate 1k episodes.

\subsection{Examples of Augmentations}
\label{app:aug_examples}

\begin{figure*}[h]
     \centering
     \includegraphics[trim={0 0 0 0},clip,width=0.85\linewidth]{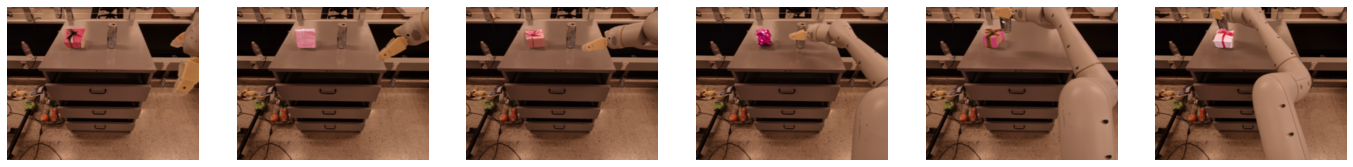}
     \includegraphics[trim={0 0 0 0},clip,width=0.85\linewidth]{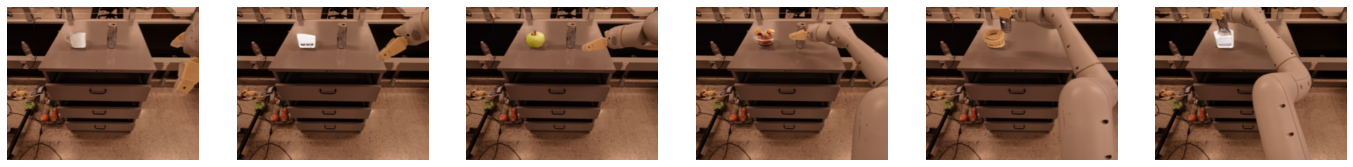}
     \includegraphics[trim={0 0 0 0},clip,width=0.85\linewidth]{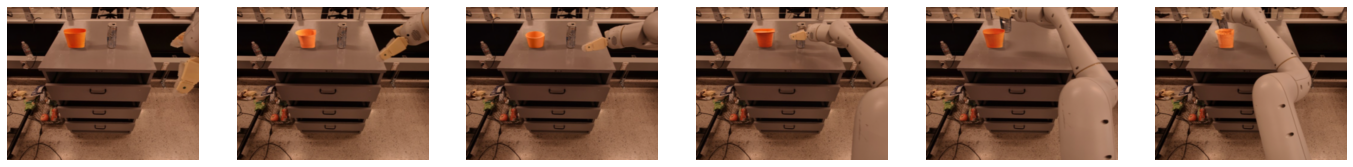}

     \caption{Augmentation Example - changing the container.
     }\label{fig:aug_example3}
\end{figure*}

We include more visualizations of augmentations generated by \algname\ in this section. In Figure~\ref{fig:aug_example3}, we show the generated episodes of \algname\ where we inpaint novel containers in the scene, which are used in the \textbf{Learning to move objects near generated novel containers} and \textbf{Learning to place objects into generated unseen containers} experiments in Section~\ref{subsec:new_skills}.

\begin{figure*}[h]
     \centering
     \includegraphics[trim={0 0 0 0},clip,width=0.85\linewidth]{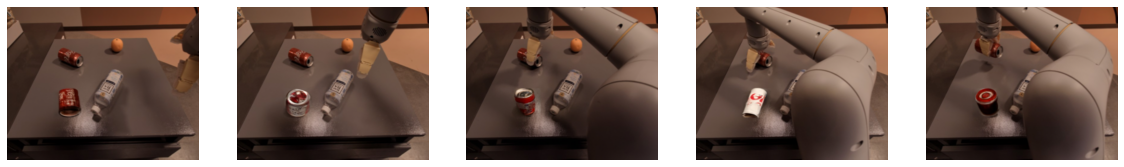}
     \includegraphics[trim={0 0 0 0},clip,width=0.85\linewidth]{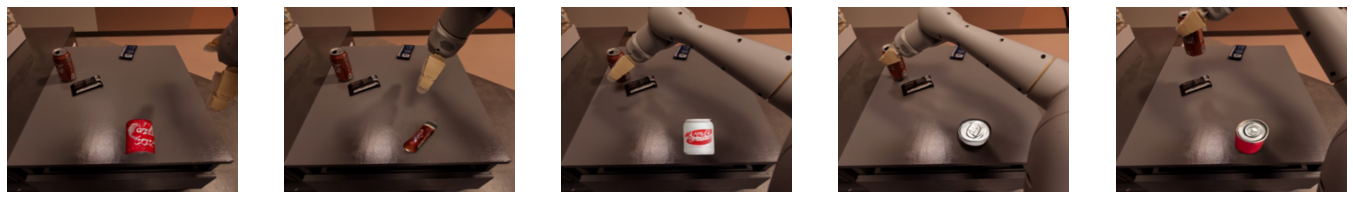}
     \caption{Augmentation Example - adding a distractor can on the table.
     }\label{fig:aug_example}
\end{figure*}

\begin{figure*}[h]
     \centering
     \includegraphics[trim={0 0 0 0},clip,width=0.85\linewidth]{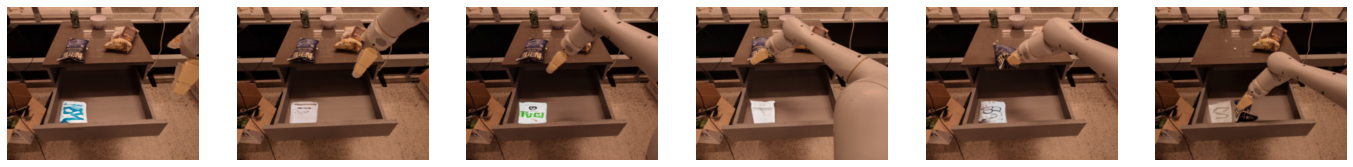}
     \includegraphics[trim={0 0 0 0},clip,width=0.85\linewidth]{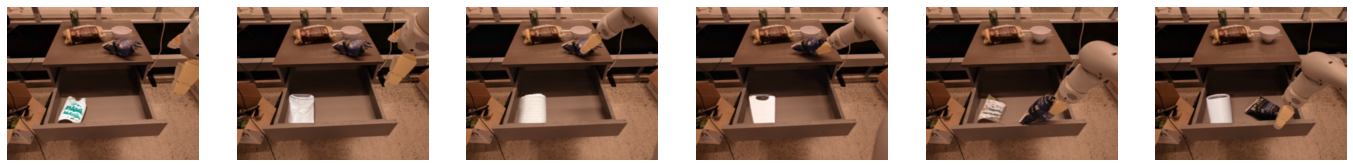}
     \caption{Augmentation Example - adding distractor objects into the drawer.
     }\label{fig:aug_example2}
\end{figure*}

In Figure~\ref{fig:aug_example} and Figure~\ref{fig:aug_example2}, we visualize augmented episodes with new distractors, e.g. cluttered coke cans on the table and chip bags in the empty open drawer. These augmentations correspond experiments conducted in Section~\ref{sec:robustify}.

\begin{figure*}[h]
     \centering
     \includegraphics[trim={0 0 0 0},clip,width=0.95\linewidth]{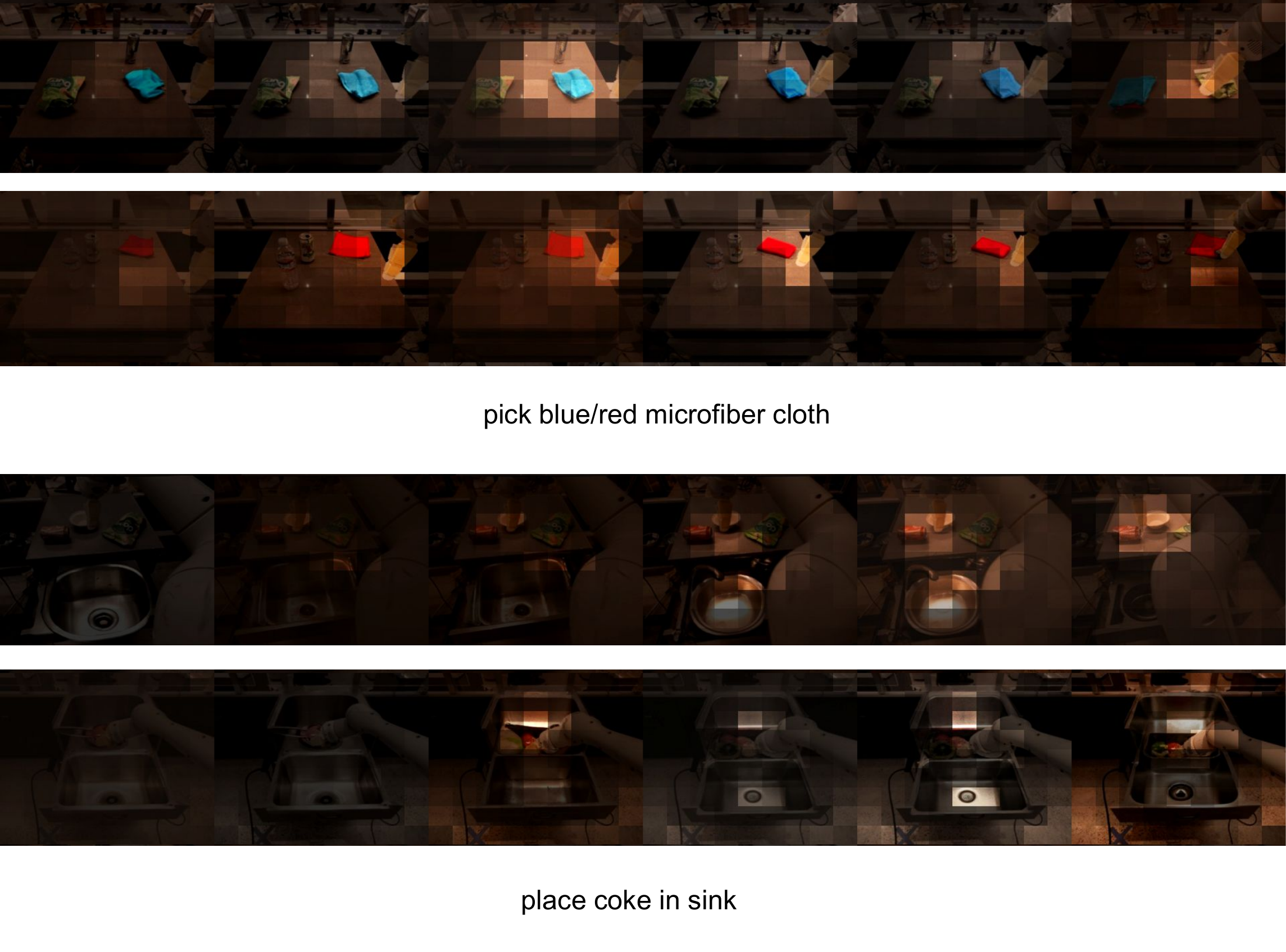}
     \caption{Visualization of some attention heads focusing on our augmented objects. This visualization is an overlay of observation and the spatial attention (bright regions means high attention).
     }\label{fig:attention_viz}
\end{figure*}

We also visualize the attention layers in RT-1 when training on our augmented data. As seen in Fig.~\ref{fig:attention_viz}, there are attention heads focusing on our augmented objects, which indicates the augmentation seem to be effective.

Overall, note that \algname\ is able generate semantically realistic novel objects and distractors in the manipulation setting. For example, \algname-generated objects typically has realistic shades on the table or the drawer, which is beneficial for training manipulation policies on top of such data.

\subsection{Failure Cases of Generated Prompts and Images}
\label{app:failure_cases}

While our LLM-assisted prompts generally work very well, we would like to note that it requires few-shot prompting to work well. In the zero-shot case, LLM would just hallucinate and output unuseful augmentation prompts. For example, if we provide the following zero-shot prompt:
\vspace{0.3em}
\noindent\fbox{\parbox{0.97\linewidth}{\scriptsize{\texttt{{%
{\color{prompt-gray}
Source task: pick coke can on a table}\\
{\color{prompt-gray}
Target task: pick coke can near a sink}\\
\query{Goal: replace the scene in the source task with the scene in the target task}\\
{\color{prompt-gray} inpainting prompt:\\
}}}}}}\\
and LLM gives the following response:\\
\vspace{0.3em}
\noindent\fbox{\parbox{0.97\linewidth}{\scriptsize{\texttt{{%
\hlcode{Pick up the coke can near the sink, replacing the one originally on the table\\}
}}}}}\\
,which is not correct. Therefore few-shot prompting is crucial in \algname.

\begin{figure}[h]
     \centering
     \includegraphics[trim={0 0 0 0},clip,width=\columnwidth]{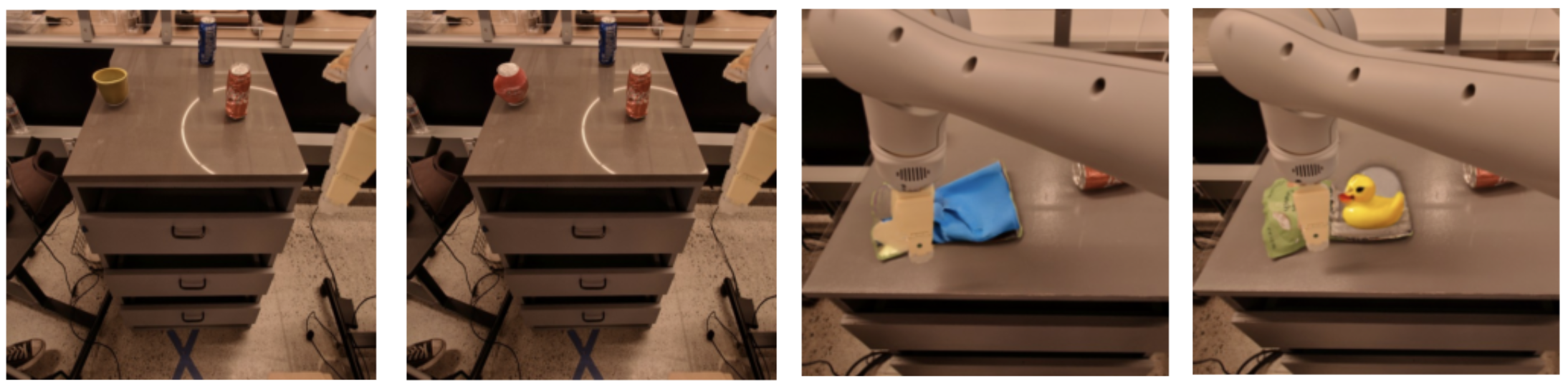}

     \caption{Failure cases of image augmentations.
     }\label{fig:failure_augment}
\end{figure}

We show the failure cases of the augmented images in Figure~\ref{fig:failure_augment}. For the two examples on the left, \algname\ is supposed to generate woven basket and glass mason jar respectively, but it fails to generate such containers and instead generate some bowl-shape containers. For the two examples on the right, \algname\ is supposed to replace the in-hand green chip bag with blue microfiber cloth and a yellow rubber duck respectively. However, as the mask of the in-hand object becomes irregular, the performance of \algname\ degrades and \algname\ is unable to generate blue microfiber cloth and the yellow rubber duck in full shape and half of the in-hand object remains as the green chip bag. We suspect that with fine-tuning \ime{} on robotic datasets that show more manipulation-related data, we can improve the generation results drastically. Note that while the generation could be suboptimal at times, our insight is that such imperfect generation can only lead to misalignment between the task instruction and images, which may not have a big negative impact on the policy results and could give extra data augmentation benefit for free. Our policy performance in Section~\ref{sec:exps} validates this insight to some degree.

\end{document}